\documentclass[letterpaper, 10 pt, conference]{ieeeconf}  

\IEEEoverridecommandlockouts                              

\usepackage{graphicx} 
\usepackage{epsfig} 
\usepackage{mathptmx} 
\usepackage{times} 
\usepackage{amsmath} 
\usepackage{amssymb}  
\usepackage{xcolor}
\usepackage{gensymb}
\usepackage{algorithm}
\usepackage{algorithmic}

\usepackage{array,multirow}
\title{\LARGE \bf
ERASE-Net: Efficient Segmentation Networks for Automotive Radar Signals
}

\author{Shihong Fang$^{1}$, Haoran Zhu$^{1}$, Devansh Bisla$^{1}$, Anna Choromanska$^{1}$, 
\\
Satish Ravindran$^{2}$, Dongyin Ren$^{2}$, Ryan Wu$^{2}$
\thanks{$^{1}$Learning Systems Laboratory, Department of Electrical and Computer Engineering,
        New York University, 370 Jay Street, NY, USA.
        {\tt\small \{sf2584, hz1922, db3484, ac5455\}@nyu.edu}}%
\thanks{$^{2}$NXP Semiconductors, San Jose, CA, USA.
        {\tt\small \{satish.ravindran, dongyin.ren, ryan.wu\}@nxp.com}}%
}

\begin{document}

\maketitle
\thispagestyle{empty}
\pagestyle{empty}

\begin{abstract}
 Among various sensors for assisted and autonomous driving systems, automotive radar has been considered as a robust and low-cost solution even in adverse weather or lighting conditions. With the recent development of radar technologies and open-sourced annotated data sets, semantic segmentation with radar signals has become very promising. However, existing methods are either computationally expensive or discard significant amounts of valuable information from raw $3$D radar signals by reducing them to $2$D planes via averaging. In this work, we introduce ERASE-Net, an Efficient RAdar SEgmentation Network to segment the raw radar signals semantically. The core of our approach is the novel detect-then-segment method for raw radar signals. It first detects the center point of each object, then extracts a compact radar signal representation, and finally performs semantic segmentation. We show that our method can achieve superior performance on radar semantic segmentation task compared to the state-of-the-art (SOTA) technique. Furthermore, our approach requires up to $20\times$ less computational resources. Finally, we show that the proposed ERASE-Net can be compressed by $\bf 40\%$ without significant loss in performance, significantly more than the SOTA network, which makes it a more promising candidate for practical automotive applications.
\end{abstract}

\section{INTRODUCTION}
\label{sec:intro}
Automotive Frequency Modulated Continuous Wave (FMCW) radar is used extensively in both Advanced Driver Assistance (ADAS) and Autonomous Driving (AD) Systems. Compared to other modalities such as cameras and LiDAR, radar has advantages such as increased robustness to adverse weather conditions and providing Doppler information of surrounding objects. These powerful features, combined with its low-cost nature, have made radar popular in the automotive industry for decades. 

Recent advancements in radar technology and the availability of open-source CARRADA~\cite{9413181} data set open up the great potential for deep learning application in radar scene understanding. 
In this work, we study the semantic segmentation problem on the radar Range-Angle-Doppler (RAD) spectrum.
This enables us to get fine-grained information, including the class of the object, its location, and the radial velocity, as well as the accurate shape. The existing SOTA  semantic segmentation algorithm~\cite{Ouaknine_2021_ICCV} used for RAD data is computationally expensive, prohibiting its implementation in edge processors. Furthermore, it removes lots of rich information in raw RAD data by reducing 3D information in the signal to 2D planes via averaging, which leads to the downgrade of its segmentation performance. Here we propose a novel DNN architecture, dubbed ERASE-Net, an Efficient RAdar SEgmentation Network to segment the raw RAD tensor semantically. Our approach takes advantage of the sparse characteristics of the RAD spectrum, extracting valuable information around the center point of the objects of interest to form a compact radar representation. Later we only conduct segmentation based on this novel radar presentation, which motivates segmenting accurately and efficiently.
We demonstrate the effectiveness of the ERASE-Net on the NXP simulated data set, where it is shown superior to the performance of the SOTA approach on radar semantic segmentation. Then we validate our approach on the public data set, CARRADA and obtain comparable results. We show that we utilize only a fraction of the compute resources of the current SOTA model. Furthermore, we show that the proposed ERASE-Net is more conducive to network pruning in memory-constrained settings, making them more suitable for real world automotive applications.

The contributions of our work can be summarized as follows: we propose a detect-then-segment radar semantic segmentation framework that detects objects' center points and then only segments point clouds in the region of interest (ROI) around the centers. This provides a new perspective for radar semantic segmentation tasks. We empirically measure ERASE-Net's segmentation performance and computational requirements, demonstrating it is the best model for real world applications.

This paper is organized as follows: Section~\ref{sec:rl} provides a background on the different radar signal representations and the review of works considering radar semantic segmentation, Section~\ref{sec:data} describes the data set, Section~\ref{sec:network} describes our approach, Section~\ref{sec:experiment} shows experimental results, and Section~\ref{sec:con} concludes the paper. 
  
\section{BACKGROUND AND RELATED WORKS}
\label{sec:rl}

\subsection{Radar signal representation}

Different authors have proposed many useful representations of the radar signal from the RAD spectrum. For example, some  works~\cite{Ouaknine_2021_ICCV, Nowruzi2020DeepOS} focused on generating the 2D RA map by applying the maximum or sum projections across the Doppler dimension. Similarly, the RD map and RA map can be created by projecting along the angle and Doppler dimensions, respectively. Radar point clouds are also widely used due to their similarity to LiDAR data. They can be generated using the constant false alarm rate (CFAR) algorithm~\cite{10.5555/561899} that detects peaks in the RAD spectrum that can be clearly discerned from their surrounding. However, a large amount of the information is discarded in this way and the resulting point clouds are much sparser compared to the LiDAR point clouds. 

\subsection{Semantic segmentation on radar data}
Recently, a few data sets containing radar data have been released. However, most of them are relatively small and rely on different radar representations. The nuScenes~\cite{nuscenes} was one of the first public data sets that  included radar. Similarly, RadarScenes~\cite{schumann_ole_2021_4559821} also only provide point clouds. Other data sets provide the radar signal in a single view. The RobotCar~\cite{RobotCarDatasetIJRR}, MulRan~\cite{1013-kim}, CRUW~\cite{wang2021rodnet}, and RADIATE~\cite{sheeny2020radiate} rely on a rotating radar but they only provide the RA map without Doppler information. On the other hand, the Zendar~\cite{Mostajabi2020HighRR} data set only uses RD maps without azimuth information. Recently released CARRADA and RADDet~\cite{DBLP:conf/crv/ZhangNL21} data sets provide the entire RAD spectrum. CARRADA uses a semi-automatic annotation approach to annotate the data set. To the best of our knowledge, this is the only data set that provides semantic annotations in both RA and RD views and therefore it is used as one of our data sets to evaluate our approach.

The existing works on radar signal segmentation mostly focus on the point cloud representation due to its similarity to LiDAR point clouds. Thus it is natural to extend the network architectures originally developed for processing the LiDAR point cloud to the radar point cloud~\cite{8455344, 10.5555/3295222.3295263, 8911477}. These approaches however have limited applicability for radar point clouds data since this data is extremely sparse.
Furthermore, thanks to the progress brought by image segmentation, the segmentation on the RA maps is explored in some recent works~\cite{kaul2020rss,DBLP:conf/vehits/NowruziKKHHLRM21, Aldera2019FastRadar}. They apply SOTA image segmentation architectures, e.g., U-Net~\cite{Ronneberger2015UNetCN} and Deeplabv3+~\cite{10.1007/978-3-030-01234-2_49}, and used effective segmentation modules such as skipped connections and atrous spatial pyramid pooling (ASPP)~\cite{10.1007/978-3-030-01234-2_49} to increase the performance.
The most recent work~\cite{Ouaknine_2021_ICCV} is the first work to utilize the raw RAD spectrum, take its projection maps as input, and produce semantic segmentation results on both RA and RD maps. Using 2D project maps as input is still computationally expensive for embedded systems of autonomous driving, and lots of information is discarded when projecting 3D RAD data to 2D planes via averaging. This apporach was evaluated on the recently-released CARRADA data set. We consider this approach as our baseline.

\section{DATA SETS}
\label{sec:data}
In this section, we explain the details of the data sets we used. Table~\ref{tbl:datastats} and ~\ref{tbl:radarpara} captures their technical difference.


\subsection{CARRADA data set}

In our experiments, we use the raw RAD tensor provided by CARRADA data set as input. The dimension of the raw tensor is $256 \times 256 \times 64$. We choose the dense mask as our annotation type. The released data set only provides dense semantic segmentation annotation in RA and RD views, but does not contain the labeling in RAD space. To obtain such annotations, we slightly modify the data pipeline provided by the authors of the CARRADA data set. To be more specific, in the last step, instead of first projecting the sparse points into RA and RD views and then using dilation to expand the region of annotations, we directly dilate the sparse points in the RAD space and then project them into RA and RD view to form the labeled RA and RD maps. We use the exact same parameters that were used in the original annotation pipeline and make sure the resulting annotated RA and RD maps are almost identical to the original ones.

\begin{table}[t]
    \centering
    \caption{data set statistics}
    \vspace{-0.1in}
    \resizebox{0.65\linewidth}{!}{
    \begin{tabular}{ccc}
    \hline Parameter & CARRADA & NXP\\
    \hline 
    Total number of instances & $78$ & $160$ \\
    Total number of frames & $12666$ & $6500$\\
    Mean number of frames per sequence & $422$ & $500$\\
    Maximum number of instances in one sequence & 2 & 16 \\
    Total number of annotated frames with instance(s) & $7193$ & $6500$\\
    \hline
    \end{tabular}}
    \label{tbl:datastats}
    \vspace{-0.1in}
\end{table}

\begin{table}[t]
    \centering
    \vspace{0.05 in}
    \caption{radar parameters}
    \vspace{-0.1in}
    \resizebox{0.6\linewidth}{!}{
    \begin{tabular}{ccc}
    \hline Parameter & CARRADA & NXP\\
    \hline 
    Frequency & $77$ GHz & $77$ GHz \\
    Sweep Bandwidth & $4$ GHz & $0.159$ GHz\\
    Maximum Range & $50$ m & $200$ m\\
    FFT Range Resolution & $0.2$ m & $0.95$ m \\
    Maximum Radial Velocity & $13.43$ m/s & $20$ m/s \\
    FFT Radial Velocity Resolution & $0.42$ m/s & $0.77$ m/s\\
    Field of View & $180\degree $ & $180\degree $\\
    FFT Angle Resolution & $0.70$ & $2.01 $\\
    Number of Chirps per Frame & $64$ & $52$\\
    Number of Samples per Chirp & $256$ & $212$\\
    \hline
    \end{tabular}}
    \vspace{-0.1in}
    \label{tbl:radarpara}
\end{table}

\begin{figure*}[ht]
  \centering
  \vspace{0.02 in}
  \includegraphics[width=0.6\textwidth]{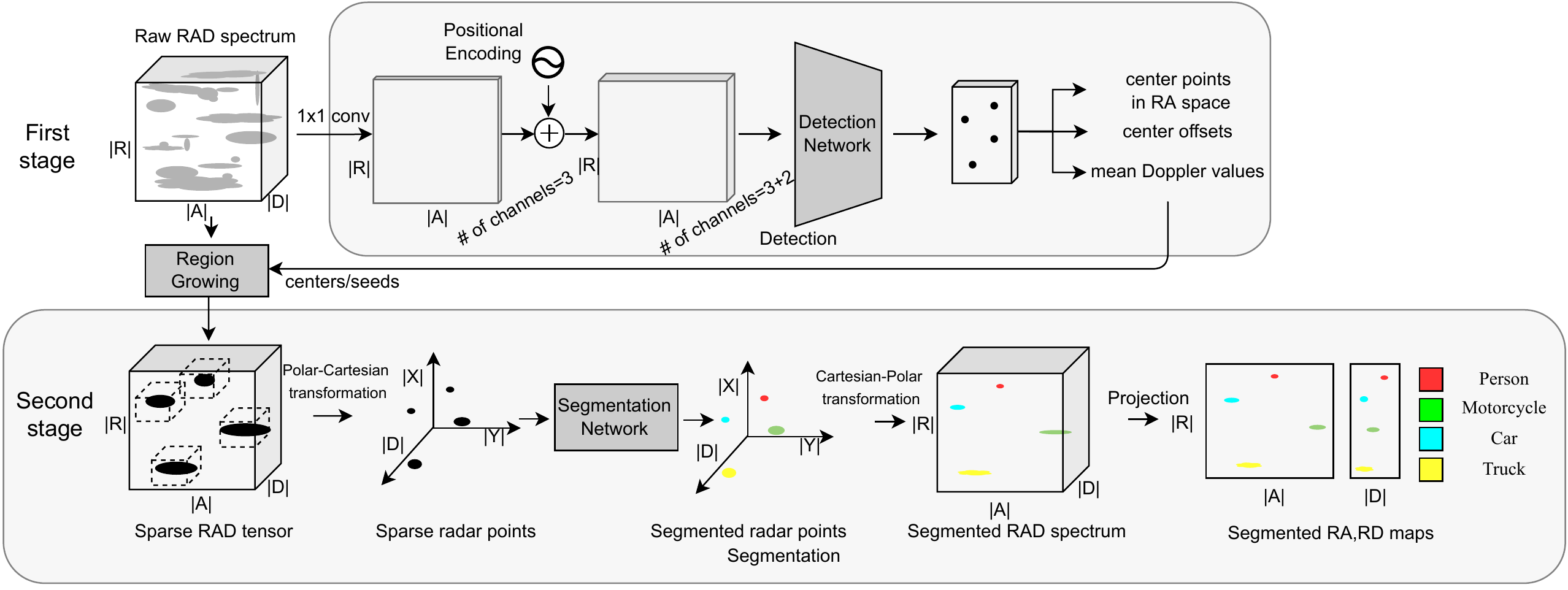}
  \makeatletter\def\@captype{figure}\makeatother
  \vspace{-0.05 in}
  \caption{The overview of ERASE-Net architecture.}
  \label{fig:erase-net}
  \vspace{-0.05 in}
\end{figure*}
\subsection{NXP simulated data set}

\begin{figure}[t]
  \centering
  \includegraphics[width=0.8\columnwidth]{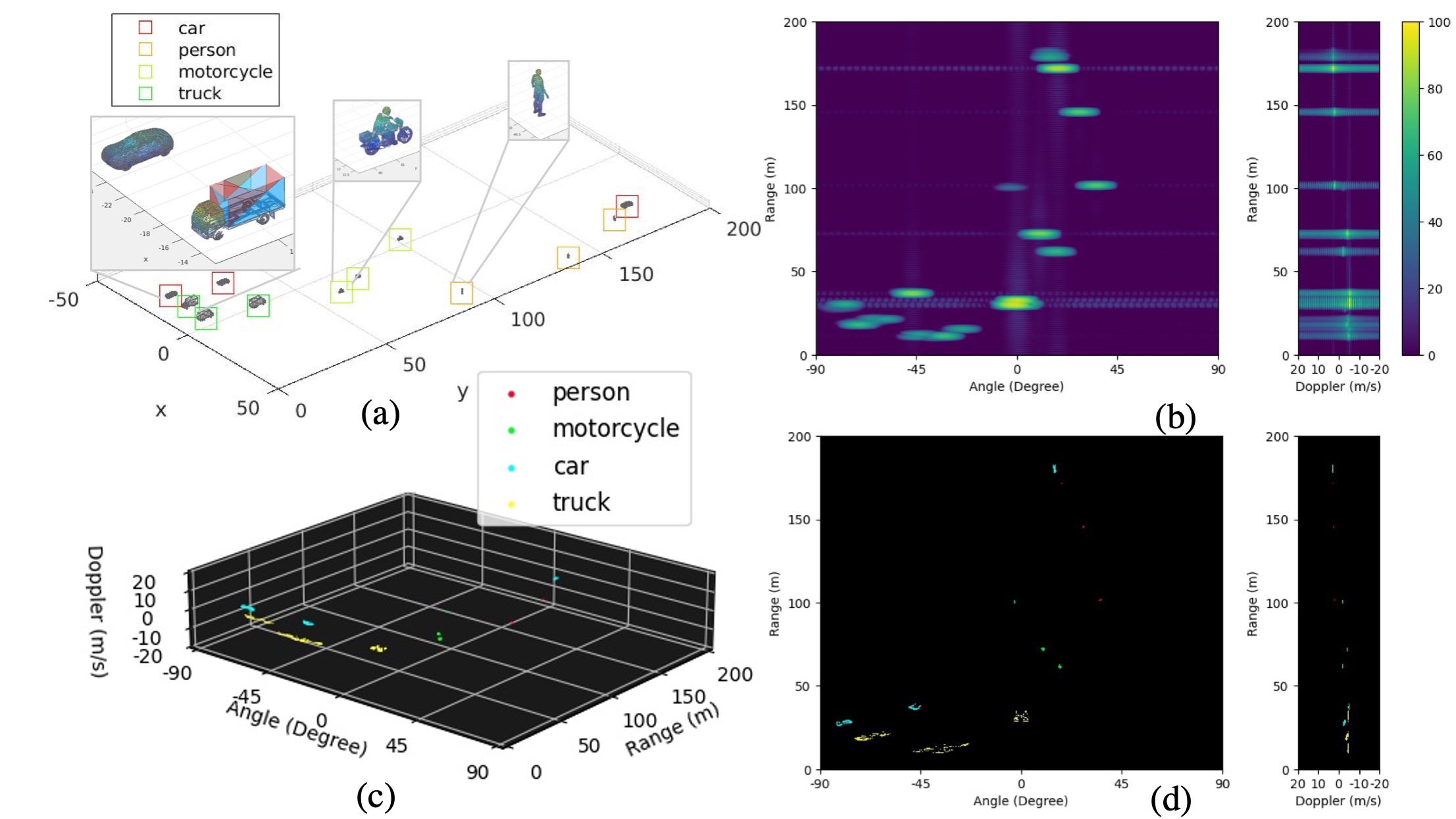}
  \vspace{-0.1 in}
  \caption{Exemplary test scene from the NXP simulated data set. (a) A screenshot of the simulated world. (b) The raw RA and RD maps. (c) The labeling in the RAD space. (d) The annotated RA and RD maps. Figure best viewed when zoomed in.}
  \label{fig:NXP}
  \vspace{-0.25 in}
\end{figure}

Since the CARRADA data set is annotated in a semi-automatic manner, we found that it suffers from mislabeling and over-labeling issues in several frames. To overcome the limitations of the CARRADA data set, mimic the real world driving/traffic scenarios, and reproduce the performance of a typical FMCW radar, NXP designed the radar system and scenario generation simulator in MATLAB. The simulated radar system parameters are presented in Table~\ref{tbl:radarpara}. 
Since we manipulated the simulated world by placing different moving objects, the labels in the simulated data set are accurate. 
We have four different types of objects in the simulator, which can be categorized as a \textit{person}, \textit{motorcycle}, \textit{truck}, or \textit{car}. and the number of objects from each category was set to be equal to avoid bias during model training. Furthermore, we created $13$ different worlds and collected $500$ frames of data from each world. The comparisons made in Table~\ref{tbl:datastats} shows the number of instances in the NXP simulated data set is more than twice the CARRADA and there are at least $12$ objects in every sequence of the recording, which makes the driving scenario more complex.
The radar data cube is obtained and is represented by the RAD tensor which has the dimensionality  $424 \times 512\times 104$. Finally, Fig.~\ref{fig:NXP} shows an exemplary scene from the data set. 

\section{ERASE-NET}
\label{sec:network}

Our proposed network utilizes ``detect-then-segment" mechanism as shown in Fig.~\ref{fig:erase-net}. It takes the entire RAD tensor as input and process it in two stages. In the first stage, we use a detection network to predict each object's center in RAD space. Then a region growing algorithm takes the predicted center points as seeds and selects the ROIs around them in the RAD spectrum. These selected regions are separated from the background, forming a compact representation of raw RAD data, and are used as the input for the segmentation network that constitutes the second stage of the data processing pipeline. Compared to the original input, the resulting tensor containing only ROIs is much more sparse. To take advantage of the sparse structure of the input radar representation, we design the final segmentation network using sparse convolutions and run semantic segmentation on the aforementioned sparse RAD tensor obtained from the first stage of data processing. By combining the background information obtained in the first stage and objects' semantics predicted in the second stage, we get the final semantic segmentation results. We discuss the details of each data processing stage below.

\subsection{First stage of data processing — detecting object centers}
The network first takes the complex RAD tensor $I \in \mathbb{C}^{|R| \times |A| \times |D|}$ as input, where $|\cdot|$ stands for the size of the given domain. 
We apply the  logarithmic transform~\cite{Ouaknine_2021_ICCV} using (\ref{eq:real}) to get the real-valued tensor $I' \in \mathbb{R}^{|R| \times |A| \times |D|}$, where RA dimensions are set as the input dimensions and the range of Doppler dimension is set to be the channel size.
\vspace{-0.01in}
\begin{equation}
\label{eq:real}
\vspace{-0.02 in}
I' = 10\log_{10}(\left|I \right|^2 + 1)
\vspace{-0.02 in}
\end{equation}

 Next, $1 \times 1$ convolution is applied to the input to compress the tensor along the Doppler axis. The compressed Doppler dimension is set to be $3$. Then in order to embed absolute positional information explicitly, two additional channels are added to the input which contain each pixel's location in Cartesian coordinates, which contain each pixel's location in Cartesian coordinates. Finally, the fused features are passed to the detection network. We adopt the CenterNet~\cite{Zhou2019ObjectsAP} detection network framework for generating preliminary objects' center predictions on the RAD spectrum. The features are first extracted using a fully convolutional encoder-decoder backbone network. We follow EfficientDet~\cite{Tan2020EfficientDetSA} and use a modified version of the EfficientDet-D0 network as the backbone. The extracted radar features are then used to predict object center points in the RA space, as well as center offset and its mean Doppler value. These values are predicted by three separate regressors, each consisting of a $3 \times 3$  convolution layer with $64$ channels followed by a $1 \times 1$ convolutional layer that forms the desired output.

The first regressor produces a center point heat map $\hat{Y} \in [0,1]^{\frac{|R|}{S} \times \frac{|A|}{S}}$, where $S$ is the output stride. The output stride downsamples the output prediction by a factor of $S$. Let $\hat{y}_{ij}$ be a score at location $(i, j)$ in the predicted heapmap $\hat{Y}$. Then as shown in Equation~\ref{eq:fg_bg}, the center points of detected targets can be separated from the background.

\vspace{-0.12in}
\begin{equation}
\label{eq:fg_bg}
\hat{y}_{ij} = 
    \begin{cases}
          1 &\text{if } \, \text{foreground} \\
          0 &\text{if } \, \text{background}. \\ 
    \end{cases}
\end{equation}
\vspace{-0.12in}

The ground truth center point $p \in \mathbb{R}^2$ of the object is generated by averaging the Cartesian coordinates of all points that belong to the object. The center point in low resolution is located at $\Tilde{p} = \lfloor\frac{p}{S}\rfloor$. We then plot all ground truth center points onto the ground truth heatmap $Y \in [0,1]^{\frac{|R|}{S} \times \frac{|A|}{S}}$ by using a Gaussian kernel centered at $\Tilde{p}$. Specifically, in the ground truth heatmap $Y$, every pixel is obtained according to:
$y_{ij} = \exp{(-\frac{(i-\Tilde{p}_i)^2}{2\sigma_{p_i}^2}-\frac{(j-\Tilde{p}_j)^2}{2\sigma_{p_j}^2})}$, where $\sigma_{p_i}$ and $\sigma_{p_j}$ are the standard deviation of the Gaussian kernel which depends on the size of the object in RA view. We determine the radius of the kernel to be $1$ and make the $\sigma_{p_i}$ to be $1/3$ of the radius following the work~\cite{law2018cornernet}. By observing the object's shape in RA view, we find that shape is like an ellipse with its long axis aligning with the angle axis and short axis aligning with the range axis. Therefore we set the $\sigma_{p_j}$ to be $1/2$ of the radius. This Gaussian kernel is rendered at each ground truth
object center.  The training objective for the first regressor uses the focal loss as shown in (\ref{eq:focal_loss}):

\vspace{-0.2in}
\begin{equation}
\label{eq:focal_loss}
\mathcal{L}_h = -\frac{1}{N} \sum_{i,j}
    \begin{cases}
          (1-\hat{y}_{ij})^{\alpha} \log(\hat{y}_{ij}) &\text{if} \, y_{ij}=1 \\
          (1-y_{ij})^{\beta} (\hat{y}_{ij})^{\alpha} \log(1-\hat{y}_{ij}) &\text{if} \, y_{ij} \neq 1, \\
     \end{cases}
\end{equation}
\vspace{-0.15in}

\noindent where N is the number of center points in an image, and $\alpha$ and $\beta$ are the hyper-parameters of the focal loss. We use $\alpha=2$ and $\beta=4$ in all our experiments.

Secondly, to recover the quantization error caused by down-sampling, the next regressor predicts the offset error  $\hat{O} \in \mathbb{R}^{\frac{|R|}{S} \times \frac{|A|}{S} \times 2}$ that can be measured as $\frac{p}{S} - \Tilde{p}$. This regressor is trained with an L1 loss:

\vspace{-0.12in}
\begin{equation}
    \mathcal{L}_o = \frac{1}{N} \sum_{p} \left|\hat{O}_{\Tilde{p}} - (\frac{p}{S} - \Tilde{p})  \right|
\label{eq:l1}
\end{equation}

The third regressor predicts the Doppler value of the center point: $\hat{D} \in \mathbb{R}^{\frac{|R|}{S} \times \frac{|A|}{S}}$, whereas the ground truth $D \in \mathbb{R}^{\frac{|R|}{S} \times \frac{|A|}{S}}$ is calculated by averaging the Doppler values of all points that are annotated as the same object. Similar to (\ref{eq:l1}), we also use L1 loss here:
\begin{equation}
    \mathcal{L}_D = \frac{1}{N} \sum_{p} \left|\hat{D} - D  \right|
    \label{eq:l1loss}
\end{equation}
The overall training objective is:

\begin{equation}
    \mathcal{L}_{det} = \mathcal{L}_h + \lambda_o \mathcal{L}_o + \lambda_D \mathcal{L}_D,
    \label{eq:overall_loss}
\end{equation}

\noindent where $\lambda_o$ and $\lambda_D$ are the hyper-parameters controlling the weight importance. We use $\lambda_o=1$ and $\lambda_D=1$. 

\subsection{Second stage of data processing — sparse radar signal representation}
Once we obtain each object's center in the first stage of data processing, we can use it to extract the ROI using simple Region Growing algorithm~\cite{adams1994seeded}. As a result we obtain a compact radar signal representation. The idea is to use the center points as seeds and search for regions that are close to the seeds in the original RAD spectrum. The regions are grown from these seed points to their 6-connected neighborhood in three dimensional RAD space depending on two criterions. In the algorithm, we set \textit{$D_{thresh}$} as our search space and \textit{$I_{thresh}$} as the lowest intensity value of the points can be considered to be included in the ROIs.

After we apply the Region Growing algorithm on the raw input tensor, we  obtain the regions that contain all possible objects. The resulting representation is much sparser compared to the original input. Then we transform the sparse tensor into a point cloud representation. We treat every bin in ROIs as a point and apply point cloud segmentation algorithm on this point cloud, specifically, for each point $\textbf{p}_k$, we first transform its location from polar to Cartesian coordinates and concatenate its location information $(x_k, y_k)$ with its Doppler information $d_k$ and intensity $i_k$ to form its features. To save computations, we design the segmentation network using sparse convolutions~\cite{3DSemanticSegmentationWithSubmanifoldSparseConvNet} as described in Section \ref{sec:experiment}, it takes the sparse RAD tensor as input and predicts the final segmentation results.

\vspace{-0.1 in}
\section{EXPERIMENTS}
\label{sec:experiment}
To validate our approach, we conduct experiments on NXP simulated data set as well as the public CARRADA data set. For the experiments on the NXP simulated data set, we use $4,500$ frames from nine different worlds for training, $1,000$ frames from two different worlds for validation, and the remaining $1,000$ frames from two distinct worlds for testing. For the experiments on CARRADA, we use the same data set splits as the ones proposed by the authors. In both experiments, we use the same experimental setup to train ERASE-Net.
We built our detection network based on EfficientDet~\cite{Tan2020EfficientDetSA}. Its main building block is the mobile inverted bottleneck MBConv~\cite{Sandler_2018_CVPR}. To make it more efficient, we reduce the number of channels of the last three MBConv Blocks by $1/3$ and decrease the number of repetitions of $4$th to $6$th MBConv Block by one. Next we train the detection model using ADAM optimizer for $300$ epochs. The initial learning rate was set
to $0.0001$ and was reduced by a factor of $10$ every $10$ epochs. In the second stage of data processing, we built our segmentation network based the Minkowski-Net~\cite{choy20194d} backbone. 
In addition, we also used SPVCNN~\cite{tang2020searching} for our segmentation network, which recently claimed to have better performance than Minkowski-Net in terms of point cloud segmentation. We follow the same design philosophy as described in~\cite{tang2020searching} and build the SPVCNN segmentation network based on the Minkowski-Net by wrapping residual Sparse Convolution blocks with the high-resolution point-based branch. We choose the voxel size to be $0.05$ meters for both networks and train them for 300 epochs with a starting learning rate of $0.24$ and cosine learning rate decay.

\subsection{Detection results}

\subsubsection{Measure}
\label{sec:det-measure}
To evaluate the performance of our detection network, we modify the standard object detection metric~\cite{Lin2014MicrosoftCC}. This modification is necessary as our detection mechanism only predicts the object centers instead of their bounding boxes. We use the distance between two objects in RAD space instead of the Intersection over Union (IoU) of the objects' bounding boxes to determine if the object has been correctly detected. For a pair of objects, we consider their distance in the Cartesian coordinates as well as in the Doppler space. We calculate their Euclidean distance in the Cartesian coordinates and the mean Doppler difference. We consider two objects to be within \textit{distance} $k$ if the Euclidean distance is within $k$ meters and at the same time their mean Doppler value difference is less than $k$ bins. 
Next we define True Positive (TP) as the case where the predicted center point and the ground truth center point are within distance $k$ from each other, whereas False Positive (FP) case corresponds to the setting, where this distance is above $k$. False Negative (FN) presents the case when the object is not detected. Then we can calculate the Precision $=\frac{TP}{TP+FP}$, Recall $=\frac{TP}{TP+FN}$ and the Average Precision (AP), which is the area under the curve (AUC) of the Precision $\times$ Recall curve. 

\subsubsection{Results}
\begin{figure}[th]
    \centering
    \includegraphics[width=0.23\textwidth]{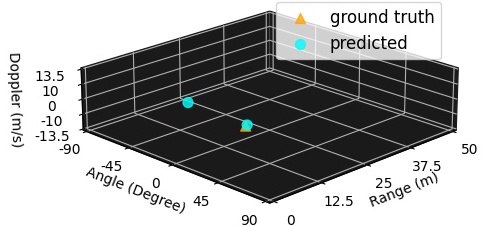}
    \includegraphics[width=0.23\textwidth]{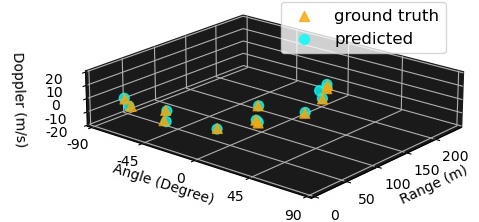}
    \vspace{-0.1 in}
    \caption{The visualization results of the objects' centers in RAD space that are predicted by our detection network. The \textbf{left} image shows the result of the network trained on CARRADA data set. The image on the \textbf{right} shows the result of the network trained on NXP data set and it is evaluated using the exemplary test scene from Fig.~\ref{fig:NXP}. Figure best viewed when zoomed in.
    }
    \vspace{-0.1 in}
    \label{fig:det}
\end{figure}

The first two regressors of the detection network predict the location of the object centers in RA space and the third estimates the mean Doppler values. These together provide the information about the predicted center point of the objects in the RAD space. Then we can calculate the distance between the predicted object centers and the ground truth center points and evaluate the performance of the network based on the metric defined in \ref{sec:det-measure}. The quantitative results on the both CARRADA and NXP test data are reported in Table~\ref{tbl:detection}. Here, by following~\cite{Lin2014MicrosoftCC}, we compute the AP when the distance ($Dist$) threshold $k$ is set to be $1, 3$ and $5$ and obtain the mean AP by averaging the AP with different $Dist$ thresholds to avoid a bias towards a specific value. As we can see from the Table, our approach detects the object center successfully by only using $2$ million parameters and extremely low GMACs. In Fig.~\ref{fig:det}, we show the detection results qualitatively for our detection networks on both CARRADA and NXP test set using the same scene. We find our detection network can accurately locate every object in the test scene. Moreover, in the CARRADA test example, our detection network can predict the location of the \textit{cyclist} even if it is not labeled in the ground truth. However, this example will be falsely counted as FP, which might explain why the mAP is lower on CARRADA than on NXP data set in Table~\ref{tbl:detection}.

\vspace{-0.1in}
\subsection{Segmentation results}
The high accuracy of our detection network provides a good anchor point for our segmentation network. We apply the Region Growing algorithm and extract the ROIs that could potentially contain the object. The objective of the Region Growing algorithm is to include all ground truth points in the search regions and at same time avoid exploring too many points to save computations. In the experiments on NXP data set, we set the \textit{$I_{thresh}$} to be half of the intensity value of each seed and in Table~\ref{tab:rg_dist} obtain the average number of points that are visited in the Region Growing algorithm using different \textit{$D_{thresh}$}. Finally, based on Table \ref{tab:rg_dist}, we select \textit{$D_{thresh}$}$=6$ in the algorithm to make sure we have a high average recall value while exploring fewer number of points. The corresponding average recall of the ROI segmentation is $96.2\%$, which means our ROIs is able to include the majority of the objects. 
We can see that our selected ROIs can successfully cover the whole ground truth points. We use similar approach to conduct the Region Growing algorithm on CARRADA data set and we obtain the average recall of $95.2\%$ by exploring $119,920$ points. To sum up, the average comparisons/GMACs done by the Region Growing algorithm for both data set are negligible compared to the computations that were made in the DNNs. 
\begin{table}[t]
\centering
    \caption{The testing results of the detection network}
    \vspace{-0.1in}
    \resizebox{\linewidth}{!}{
    \begin{tabular}{|c|c|c|c|c|c|c|}
     \hline
     \multirow{2}{*}{Backbone} & \multirow{2}{*}{Param.} (M) & \multirow{2}{*}{GMACs} &
     \multicolumn{4}{c|}{AP}\\
     \cline{4-7}
     & & & Dist-1  & Dist-3 & Dist-5 & \textbf{mAP}\\
     \hline
     \multicolumn{1}{{|c|}}{\begin{tabular}[{|c|}]{@{}c@{}}EfficientDet-D0-smaller\\ (CARRADA)\end{tabular}} & 2.08 & 0.49 & 0.19 & 0.61 & 0.76 & 0.52 \\
     \hline
     \multicolumn{1}{{|c|}}{\begin{tabular}[{|c|}]{@{}c@{}}EfficientDet-D0-smaller\\ (NXP)\end{tabular}} & 2.08 & 1.46 & 0.67 & 0.933 & 0.958 & 0.853 \\
     \hline
    \end{tabular}
    }
    \vspace{-0.05in}
    \label{tbl:detection}
\end{table}
\begin{table}[t]
\vspace{-0.05in}
\centering
\setlength\tabcolsep{1pt}
\renewcommand{\arraystretch}{1}
\caption{Average recall values for different values of Distance threshold in Region Growing algorithm on NXP val set}
\vspace{-0.1in}
\begin{tabular}{|c|c|c|c|c|c|c|}
\hline
Distance threshold & 3  & 4 & 5 & 6 & 7 & 8  \\ \hline
\begin{tabular}[c]{@{}l@{}}\ Average \# of \\ visited points\end{tabular}
 & 20,731 & 41,579 & 66,527 & 94,504 & 126,127 & 160,323 \\ \hline
 Average recall   
 & 0.901 & 0.946 & 0.958 & 0.962 & 0.965 & 0.967 \\ \hline
\end{tabular}
\vspace{-0.2in}
\label{tab:rg_dist}
\end{table}

\begin{figure*}[ht!]
    \centering
    \vspace{0.08 in}
    \includegraphics[width=0.9\linewidth]{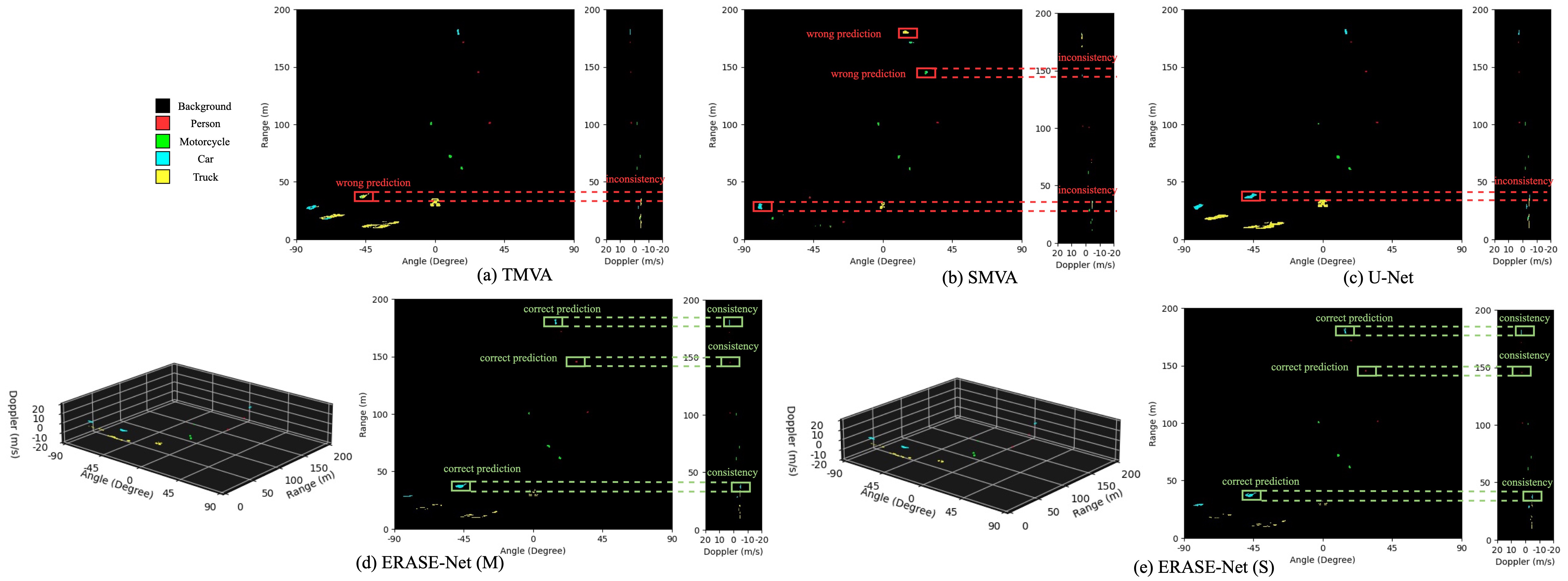}
    \vspace{-0.1in}
    \caption{Qualitative comparisons between ERASE-Nets (d, e) and competitor methods (a-c) when they are tested on the NXP test example shown in Fig.~\ref{fig:NXP}. The ground truth in RAD, RA and RD are also shown in Fig.~\ref{fig:NXP}. The box in red shows the wrong segmentation results predicted by the network whereas the box in green indicates the correct predictions. The red dotted lines shows the labelling inconsistency between RA and RD maps whereas green lines shows the consistency. Figure best viewed when zoomed in.}
     \vspace{-0.15in}
    \label{fig:seg-results}
\end{figure*}

Finally, we evaluate the performance of our segmentation networks using both Minkowski-Net and SPVCNN on both CARRADA and NXP test set. For the baseline model, we first consider the classic image segmentation network U-Net~\cite{UNet2015Ronneberger} and train two separate U-Nets using projected RA and RD maps to predict the segmentation mask in RA and RD view respectively. We also train the Temporal Multi-View network with ASPP modules (TMVA-Net)~\cite{Ouaknine_2021_ICCV}, which is the SOTA method in radar segmentation, and also our baseline for comparison. Its variant, Single-frame Multi-View network with ASPP modules (SMVA-Net), is also considered, it relies on single frame for training and testing, just like our network. ERASE-Net (M) (our approach using Minkowski network as the segmentation network backbone) and ERASE-Net (S) (our approach using SPVCNN network as the segmentation network backbone) share the same detection network. The parameter size and GMACs of our ERASE-Nets can be calculated by combining the parameter size and GMACs of both of the detection network and the segmentation network. The quantitative results are shown in Table ~\ref{tbl:seg}. All the models are able to predict the radar segmentation in both RA and RD view simultaneously except for U-Net, which is trained to predict single map only. As we can see from the top table, ERASE-Net can achieve higher mIoU than all baseline models while using fewer parameters when evaluated on the NXP data set. The computational cost of our models are significantly smaller. We achieve more than $20\times$ reduction in computations compared to the other techniques. The qualitative result is shown in Fig.~\ref{fig:seg-results}. In all our comparators, the segmentation results in RA and RD view are predicted by two separate networks/subnetworks, therefore the consistency between the labels in RA and RD maps are not guaranteed. In other words, we can clearly see from Fig.~\ref{fig:seg-results}(a-c), in the same range, there are objects only predicted in the RA maps but not labeled in the RD maps. However, our ERASE-Net makes predictions in the three dimensional RAD space and the predicted labels in RA and RD space are both accurate and consistent. In addition, we also compare the performance of ERASE-Net with the baselines in Table.~\ref{tbl:seg}(bottom), we find our approach have comparable predictions in RD and better performance in RA than the baseline models. Similar to the results on NXP data, the computational reductions are significant. 

\begin{table}[ht!]
\vspace{-0.05in}
    \centering
    \caption{Segmentation results of different models for NXP \textbf{(top)} data set and  CARRADA \textbf{(bottom)}}
    \vspace{-0.1in}
    \resizebox{\linewidth}{!}{
    \begin{tabular}{|c|c|c|c|c|c|c|c|c|c|}
     \hline
     \multirow{2}{*}{Output View} & \multirow{2}{*}{Model} & \multirow{2}{*}{Param.} (M) & \multirow{2}{*}{GMACs} &
     \multicolumn{6}{c|}{IoU(\%)(testing)}\\
     \cline{5-10}
     &  & & & Bkg. & Person & Motorcycle & Car & Truck& \textbf{mIoU}\\
     \hline
     RD 
    & U-Net & 31.04 & 36.54 & 99.91 & 73.79 & \textbf{79.28} & 36.78 & 57.91 & 69.53\\ 
    & TMVA-Net & 6.21 & 302.18 & 99.92 & 77.37 & 79.05 & 45.83 & 56.06 & 71.65 \\
    & SMVA-Net & 5.54 & 296.33 & 99.92 & 81.08 & 72.09 & 47.84 & \textbf{60.66} & 72.32 \\
   
    & \textbf{ERASE-Net (M)} & \textbf{2.93+2.08} & \textbf{12.0+1.46}   & 99.93 & 85.42 & 71.61 & 51.61 & 59.94 & \textbf{73.7} \\
	& \textbf{ERASE-Net (S)} & 2.93+2.08 & 13.0+1.46  & \textbf{99.93} & \textbf{86.14} & 69.93 & \textbf{52.8} & 58.14 & 73.39 \\

     \hline
     RA
     & baseline-U-Net& 31.04 & 181.32 & 99.84 & 55.96 & \textbf{54.47} & 41.20 & 33.86 & 57.06\\ 
     & baseline-TMVA-Net & 6.21 & 302.18 & 99.86 & 59.03 & 51.44 & \textbf{42.4} & 36.34 & 57.81 \\
     & baseline-SMVA-Net & 5.54 & 296.33 &  99.86 & 62.05 & 46.06 & 34.58 & 36.46 & 55.80 \\
     & \textbf{ERASE-Net (M)} & \textbf{2.93+2.08} & \textbf{12.0+1.46}  & 99.9 & 66.46 & 49.36 & 31.79 & 37.59 & 57.02 \\
     & \textbf{ERASE-Net (S)} & 2.93+2.08 & 13.0+1.46 & \textbf{99.91} & \textbf{68.65} & 48.84 & 35.46 & \textbf{38.42} & \textbf{58.26} \\
     \hline
    \end{tabular}}
    \newline
    \vspace{0.20 in}
    \newline
    \resizebox{\linewidth}{!}{
    \begin{tabular}{|c|c|c|c|c|c|c|c|c|}
     \hline
     \multirow{2}{*}{Output View} & \multirow{2}{*}{Model} & \multirow{2}{*}{Param.} (M) & \multirow{2}{*}{GMACs} &
     \multicolumn{5}{c|}{IoU(\%)(testing)}\\
     \cline{5-9}
     &  & & & Bkg. & Person & Cyclist & Car & \textbf{mIoU}\\
     \hline
     RD 
     & U-Net & 31.04 & 54.84 & \textbf{99.74} & \textbf{53.92} & 23.67 & \textbf{54.72} & \textbf{58.01}\\
     & TMVA-Net & 5.63 & 102.35 & 99.68 & 51.68 & 21.05 & 55.71 & 57.03\\
     
     & SMVA-Net & 4.74 & 72.35 & 99.67 & 50.50 & 22.89 & 52.49 & 56.39\\
     & \textbf{ERASE-Net (M)} & \textbf{2.93+2.08} & \textbf{8.41+0.49}  & 99.68 & 45.47 & \textbf{24.47} & 44.65 & 53.57 \\
     & \textbf{ERASE-Net (S)} & 2.93+2.08 & 9.3+0.49  & 99.68 & 36.89 & 18.5 & 49.95 & 51.26 \\
     \hline
     RA
     & U-Net & 31.04 & 13.71 & 99.80 & 2.15 & 1.64 & 19.55 & 30.42\\
     & TMVA-Net & 5.63 & 102.35 & 99.75 & 18.60 & 6.29 & 26.79 & 37.86\\
     & SMVA-Net & 4.74 & 72.35 & 99.73 & 18.12 & 8.88 & 25.96 & 38.17 \\
     & \textbf{ERASE-Net (M)} & \textbf{2.93+2.08} & \textbf{8.41+0.49}  & \textbf{99.86} & \textbf{31.99} & \textbf{16} & 27.94 & \textbf{43.95} \\
     & \textbf{ERASE-Net (S)} & 2.93+2.08 & 9.3+0.49 & \textbf{99.86} & 26.7 & 12.47 & \textbf{32.42} & 42.86 \\
     \hline
    \end{tabular}
    }
    \label{tbl:seg}
\end{table}
\subsection{Network Compression}
\label{subsec:network_compression}
\begin{figure}[ht]
    \includegraphics[width=0.23\textwidth]{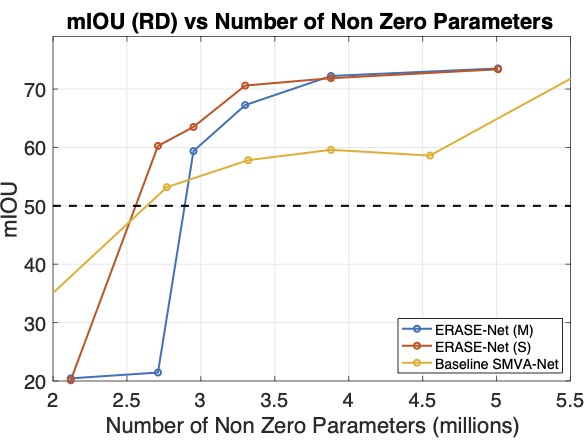}
    \includegraphics[width=0.23\textwidth]{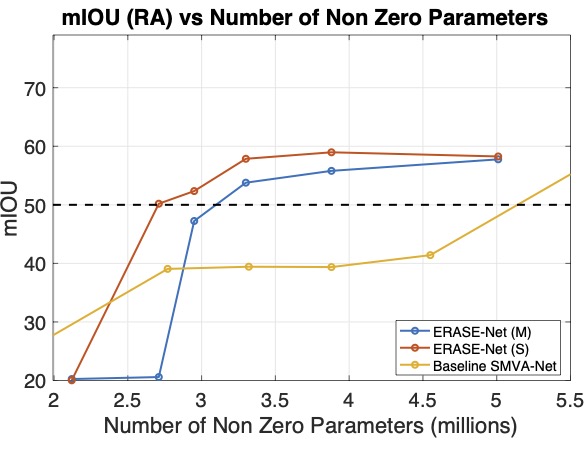}
    \vspace{-0.05in}
    \caption{Mean IOU for RD (\textbf{left}) and RA (\textbf{right}) view vs number of non-zero parameters for baseline SMVA-Net, ERASE-Net (M) and ERASE-Net (S) on NXP simulated test data set. Note that we set mean IOU value of 50 as an acceptable value for practical applications.}
    \vspace{-0.2in}
    \label{fig:mior_non_zero_params}
\end{figure}

In order to prune radar segmentation networks, we utilize network slimming technique previously presented in \cite{Liu2017learning}. In particular, we associate a scaling factor (parameter $\gamma$ of the batch normalization layer) with each channel of the convolution filter and include a sparsity loss ($\mathcal{L}_{s}$) on these parameters in the loss function. The loss function takes the form of $\mathcal{L}_{det} = \mathcal{L}_h + \lambda_o \mathcal{L}_o + \lambda_D \mathcal{L}_D + \lambda_{s}\mathcal{L}_{s}$. The value of hyperparameter $\lambda_{s}$ was set to $1e^{-4}$ for all our experiments. The networks were retrained and the filters with small value of aforementioned scaling parameters ($\gamma$) were considered unimportant and removed from the network. The networks were further fine tuned for $50$ epochs. During the fine tuning, learning rate for baseline networks was set to $1e^{-5}$ and learning rate for ERASE-Net (M) and ERASE-Net (S) was set to $1e^{-3}$. We utilize the detection network pruned by $\times0.4$ as the backbone for pruning ERASE-Net (M) and ERASE-Net (S) networks. All other hyperparameters were set to values presented in Section \ref{sec:experiment}. Table~\ref{tab:prune_detection_net} shows the compression results for detection network when the distance ($Dist$) threshold $k$ is set to be $1, 3$ and $5$.
Figure~\ref{fig:mior_non_zero_params} shows mean IoU vs number of non zero parameters for various segmentation networks. The proposed architectures outperform the baseline for wide range of sizes of compressed models.

\begin{table}[ht]
\centering
\setlength\tabcolsep{1pt}
\renewcommand{\arraystretch}{1}
\caption{Detection results on NXP simulated test data set for different levels of pruning}
\vspace{-0.1in}
\resizebox{0.7\linewidth}{!}{
\begin{tabular}{|c|c|c|c|c|c|c|}
\hline
Model  & Comp & \#Params(M) & Dist-1 & Dist-3 & Dist-5 & mAP  \\ \hline
\multirow{5}{*}{EfficientDet-D0-smaller} & 0 & 2.08 & 0.67 & 0.933 & 0.958 & 0.853 \\ \cline{2-7}
 & x0.1 & 1.87 & 0.60 & 0.91 & 0.95 & 0.82 \\ \cline{2-7}
 & x0.3 & 1.45 & 0.64 & 0.91 & 0.95 & 0.83 \\ \cline{2-7}
 & x0.4 & 1.25 & 0.62 & 0.93 & 0.96 & 0.84 \\ \cline{2-7}
 & x0.6 & 0.83 & 0.52 & 0.90 & 0.94 & 0.79 \\ \cline{2-7}
 & x0.7 & 0.62 &  0.0 & 0.5 & 0.38 & 0.13 \\ \hline
\end{tabular}
}
\vspace{-0.05in}
\label{tab:prune_detection_net}
\end{table}

\section{CONCLUSIONS}
\label{sec:con}
In this work, we present ERASE-Net, an efficient network for radar segmentation. Our approach is based on a ``detect-then-segment" mechanism that first detects the object centers to extract rich information from raw RAD data and forms a compact radar representation, and then performs semantic segmentation efficiently by leveraging this novel representation. We conduct our experiments on NXP simulated data set and CARRADA data and show that our ERASE-Net is efficient and powerful. The experiments on NXP data set shows it is able to outperforms the SOTA approaches with more than 20$\times$ computational reduction. To better fit industrial requirements, we finally apply network compression on the models and find our model can achieve even better performance than the baselines in the literature, making it the best model to be applied in the real-world application.




\vspace{-0.1in}
\section*{ACKNOWLEDGMENT}
We acknowledge the award from NXP Semiconductors that sponsors the research work presented in this paper.




\bibliographystyle{IEEEtran}
\bibliography{IEEEexample}

\end{document}